\documentclass[twoside,11pt]{article}

\pdfoutput=1

\usepackage{jmlr2e}

\usepackage{times}
\usepackage[colorlinks=true, urlcolor=black, linkcolor=black, citecolor = black]{hyperref}

\usepackage{graphicx} 

\usepackage{subfigure} 
\usepackage{tikz}

\usepackage{verbatim}

\usepackage{natbib}

\usepackage{amsmath}

\usepackage{relsize}

\usepackage{booktabs}

\usepackage{multirow}

\usepackage{color}
\usepackage{url}

\usepackage{xspace} 

\usepackage{comment}

\usepackage{bm}
\usepackage{algorithm}

\usepackage{xr}

\usepackage{amsopn}

\DeclareMathOperator{\diag}{diag} 
\DeclareMathOperator{\vect}{vec}  
\def\trans{^{\sf T}}              
\newcommand{\dif}{\mathrm{d}}     

\newcommand{\eyeMatrix}[1]{\bI_{\mathsmaller{#1\times#1}}}

\newcommand{\oneVector}[1]{\mathbf{1}_{\mathsmaller{#1}}}

\def\bX{\mathbf{X}}
\def\bv{\mathbf{v}}

\def\bx{\mathbf{x}}
\def\bz{\mathbf{z}}

\def\bL{\mathbf{L}}
\def\by{\mathbf{y}}
\def\bW{\mathbf{W}}
\def\bw{\mathbf{w}}

\def\bK{\mathbf{K}}
\def\bk{\mathbf{k}}
\def\bT{\mathbf{T}}
\def\bI{\mathbf{I}}
\def\bt{\mathbf{t}}

\def\bD{\mathbf{D}}
\def\bI{\mathbf{I}}
\def\bA{\mathbf{A}}
\def\bB{\mathbf{B}}
\def\bC{\mathbf{C}}
\def\bR{\mathbf{R}}
\def\bM{\mathbf{M}}
\def\bG{\mathbf{G}}
\def\bS{\mathbf{S}}
\def\be{\mathbf{e}}
\def\bSigma{\boldsymbol{\Sigma}}
\def\bsigma{\boldsymbol{\sigma}}
\def\bOmega{\boldsymbol{\Omega}}

\def\bkappa{\boldsymbol\kappa}

\def\wfunc{\boldsymbol\omega}

\def\wHB{\bar{\bw}} 

\def\subNew{\star}
\def\newW{\bw_{\!\subNew}}
\def\newX{\bx_{\subNew}}
\def\newY{y_{\subNew}}
\def\newT{\bt_{\subNew}}
\def\newZ{z_{\subNew}}

\definecolor{mygreen}{rgb}{0,0.25,0}

\def\eg{{\em e.g., }}

\def\ourModel{\textsl{isoVCM}\xspace} 
\def\ourModelLinear{\textsl{isoVCM}$^{\textsl{lin}}$\xspace} 
\def\ourModelRBF{\textsl{isoVCM}$^{\textsl{mat}}$\xspace} 

\def\baselineGPNaiveLinear{\textsl{GP}$_{\bx}^{\textsl{lin}}$\xspace} %
\def\baselineGPTimeGeoLinear{\textsl{GP}$_{\bx,\bt}^{\textsl{lin}}$\xspace} %

\def\baselineGPNaiveRBF{\textsl{GP}$_{\bx}^{\textsl{mat}}$\xspace} %
\def\baselineGPTimeGeoRBF{\textsl{GP}$_{\bx,\bt}^{\textsl{mat}}$\xspace} %

\def\baselineFanLinear{\textsl{Fan \& Zhang$^\textsl{lin}$}\xspace}
\def\baselineFanRBF{\textsl{Fan \& Zhang$^\textsl{mat}$}\xspace}

\begin{document}

\title{Varying-Coefficient Models with Isotropic Gaussian Process Priors}

\author{\name{Matthias Bussas} \email{matthias.bussas.14@ucl.ac.uk}\\   
			 \addr University College London, Department of Statistical Science\\
		   London WC1E 6BT, United Kingdom\\
       \AND
       \name Christoph Sawade \email christoph@soundcloud.com \\
       \addr SoundCloud Ltd. \\
       Greifswalder Str. 212-213, 10405 Berlin, Germany\\
       \AND
       \name Tobias Scheffer \email scheffer@cs.uni-potsdam.de \\
       \addr University of Potsdam, Department of Computer Science\\
       August-Bebel-Strasse 89, 14482 Potsdam, Germany\\
       \AND
       \name Niels Landwehr \email landwehr@cs.uni-potsdam.de \\
       \addr University of Potsdam, Department of Computer Science\\
       August-Bebel-Strasse 89, 14482 Potsdam, Germany\\
}
\editor{}

\maketitle

\begin{abstract}
We study learning problems in which the conditional distribution of the output given the input 
varies as a function of additional task variables. In varying-coefficient models with Gaussian process priors, a Gaussian process generates the functional relationship between the task variables and the parameters of this conditional.
Varying-coefficient models subsume hierarchical Bayesian multitask models, but also generalizations
in which the conditional varies continuously, for instance, in time or space. However,
Bayesian inference in varying-coefficient models is generally intractable. 
We show that inference for varying-coefficient models with isotropic Gaussian process priors resolves to standard inference for a Gaussian process that can be solved efficiently. 
MAP inference in this model resolves to multitask learning using task and instance kernels, and 
inference for hierarchical Bayesian multitask models can be carried out efficiently using graph-Laplacian kernels.
We report on experiments for geospatial prediction.
\end{abstract} 

\section{Introduction}
\label{sec:introduction}

In standard settings of learning from independent and identically distributed \emph{(iid)} data, labels $y$ of training and test instances $\bx$ are drawn independently and are governed by 
a fixed conditional distribution $p(y|\bx)$.
A great variety of problem settings relax this assumption; they are widely referred to as \emph{transfer learning}. 
We study a general transfer learning setting in which the conditional $p(y|\bx)$ is assumed to vary as a function of additional 
observable variables $\bt$. The variables $\bt$ can identify a specific domain that an observation was drawn from (as in \emph{multitask learning}), 
or can be continuous attributes that describe, for instance, the time or location at which an observation was made (sometimes 
called \emph{concept drift}).

A natural model for this  setting is to assume a  conditional $p(y|\bx;\bw)$ with parameters
$\bw$ that vary with $\bt$. Such models are known as \emph{varying-coefficient models} 
\citep[\emph{e.g.,}][]{hastie1993varying, gelfand2003spatial}.
In {\em iid} learning, it is common to assume an isotropic Gaussian prior $p(\bw)$ over  model parameters. When the parameters vary as a function of a task variable $\bt$, it is natural to instead assume a Gaussian process (GP) prior over functions that map values of $\bt$ to values of $\bw$.
A Gaussian process implements a prior $p(\wfunc)$ over functions $\wfunc: \mathcal{T} \rightarrow \mathbb{R}^m$ that
couple parameters $\bw \in \mathbb{R}^m$ for different values of $\bt \in \mathcal{T}$ and 
make it possible to generalize over different domains, time, or space.
While this model allows to extend Bayesian inference naturally to a variety of transfer learning problems, 
inference in these varying-coefficient models for large problems is 
often impractical:
It involves Kronecker products that result in matrices of size $nm \times nm$, with $n$ the number of instances and $m$ the number of attributes \citep{gelfand2003spatial,wheeler2006bayesian}.

Alternatively, varying-coefficient models can be derived in a regularized risk minimization framework.
Such models infer point estimates of parameters $\bw$ for different observed values of $\bt$ under some model that expresses how $\bw$ changes smoothly with $\bt$~\citep{fan2008statistical}.
At test time, point estimates of $\bw$ are required for all $\bt$ observed at the test data points. 
This is again computationally challenging because typically a separate optimization problem needs to be solved for each test instance.
Most prominent are estimation techniques based on kernel-local smoothing~\citep{fan2008statistical,wu2000kernel,fan2005profile}. 

In this paper, we explore Bayesian varying-coefficient models in conjunction with isotropic Gaussian process priors. An isotropic prior encodes the assumption that elements of the vector of model parameters are generated independently of one another; isotropic GP priors are in direct analogy to isotropic Gaussian priors that are widely used in {\em iid} learning. 
Our main theoretical result is that Bayesian inference in varying-coefficient models with isotropic Gaussian process priors is equal to Bayesian inference in a standard Gaussian process with a specific product kernel. 
The main practical implication of this result is that inference for varying-coefficient models becomes practical by using standard GP tools.
Our theoretical result also leads to insights regarding existing transfer learning methods: First, we identify the exact modeling assumptions under which Bayesian inference amounts to multitask learning using a Gaussian process with task kernels and instance kernels \citep{bonilla2007kernel}.
Secondly, we show that hierarchical Bayesian multitask models~\citep[\emph{e.g.},][]{gelman1995bayesian,finkel2009hierarchical} can be represented 
as Gaussian process priors; inference then resolves to inference in standard Gaussian processes with multitask kernels based on graph 
Laplacians~\mbox{\citep{evgeniou2005learning,alvarez2011kernels}}.

Our main empirical result is that varying-coefficient models with GP priors are an effective and efficient model for prediction problems in which the conditional distribution of the output given the input varies in time and geographical location. In our experiments, varying coefficient models outperform reference models for the problems of predicting rents and real-estate prices.

The paper
is structured as follows. Section~\ref{sec:model} describes the problem setting and the varying-coefficient model.
Section~\ref{sec:inference} studies Bayesian inference and presents our main results.
Section~\ref{sec:exp} presents experiments on prediction of real estate sales prices and monthly rents; Section~\ref{sec:conc} discusses related work and concludes.

\section{Problem Setting and Model}
\label{sec:model}

\begin{figure}[t]
	\begin{center}
	\vspace{4mm}
		\includegraphics[width=0.28\linewidth]{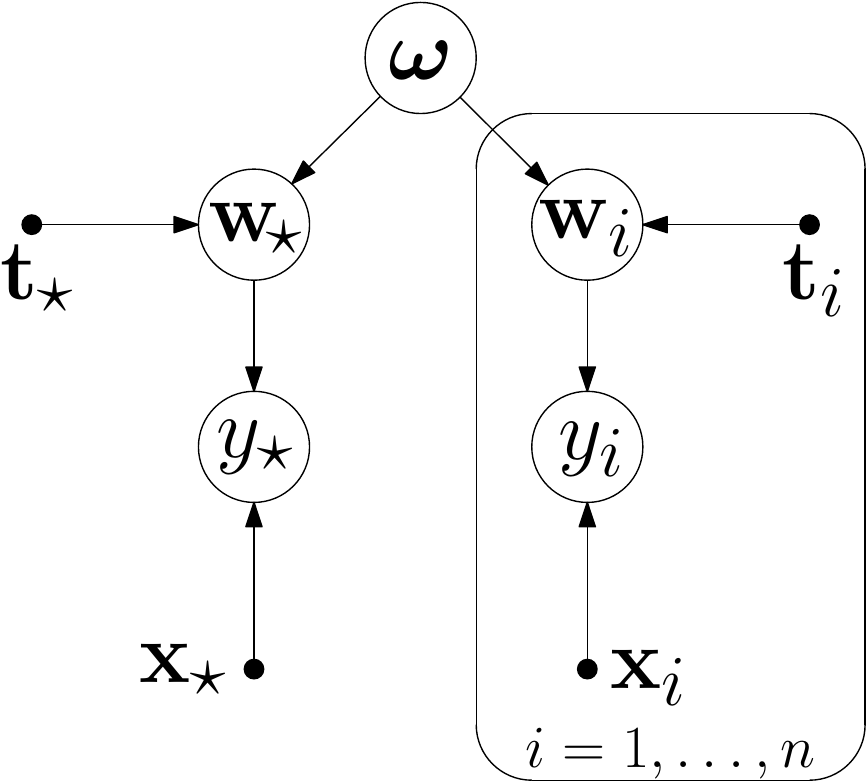}
	\vspace{3mm}
		\caption{
		Generative process of the varying-coefficient model described in Section~\ref{sec:model}. Variables $\newX, \newY, \newT, \newW$ denote the feature vector,
		label, task variable, and parameterization for a novel test instance.
		}
		\label{fig:graphical_model}
	\end{center}
\end{figure}

This section defines a generative process which models a wide class of applications that are characterized by a conditional distribution $p(y|\bx,\bw)$ whose parameterization $\bw$ varies as a function of additional variables $\bt$. 
Figure~\ref{fig:graphical_model} shows a plate representation of the  model.

A fixed set of instances $\bx_1,\dots,\bx_n$ with $\bx_i\in\mathcal{X}\subseteq\mathbb{R}^m$ is observable, along with values \mbox{$\bt_1,\dots,\bt_n \in \mathcal{T}$} of a \emph{task variable}. 
The process starts by drawing a function $\wfunc:\mathcal{T}\to \mathbb{R}^m$ according to a prior  $p(\wfunc)$.
The function $\wfunc$ associates any task variable $\bt \in \mathcal{T}$ with a corresponding parameter vector $\wfunc(\bt) \in \mathbb{R}^m$ 
that defines the conditional distribution $p(y|\bx,\wfunc(\bt))$ for task $\bt \in \mathcal{T}$.
The domain $\mathcal{T}$ of the task variable depends on the application at hand. In the simplest case of {\em multitask learning}, $\mathcal{T}=\{1,\dots,k\}$ 
is a set of task identifiers. 
In hierarchical Bayesian multitask models,
a tree $\mathcal{G} = (\mathcal{T},\bA)$ over the tasks $\mathcal{T} = \{1,\dots,k\}$ 
reflects how tasks are related; we represent this tree by its adjacency matrix $\bA \in \mathbb{R}^{k \times k}$. 
We also study the setting of {\em concept drift} or {\em non-stationary learning} in which the conditional distribution of $y$ given $\bx$ varies smoothly in the task variables $\bt$ that can, for instance, comprise time or space. In this case, $\mathcal{T} \subset \mathbb{R}^d$ is a continuous-valued space. 

We model $p(\wfunc)$ using a zero-mean Gaussian process 
\begin{equation}
	\label{eq:gp}
	\wfunc \sim \mathcal{GP}(\mathbf{0},\bkappa)
\end{equation}
that generates vector-valued functions \mbox{$\wfunc:\mathcal{T}\to \mathbb{R}^m$}. 
The process is specified by 
a matrix-valued kernel function \mbox{$\bkappa: \mathcal{T}\times \mathcal{T} \rightarrow \mathbb{R}^{m \times m}$} that reflects closeness in $\mathcal{T}$. 
Here, $\bkappa(\bt,\bt') \in \mathbb{R}^{m\times m}$ is the matrix of covariances between components of the
vectors $\wfunc(\bt)$ and $\wfunc(\bt')$ for $\bt, \bt' \in  \mathcal{T}$.
We assume that the kernel function $\bkappa$ is isotropic; that is, $\bkappa(\bt,\bt') = k_\mathcal{T}(\bt,\bt')\eyeMatrix{m}$ for a positive semidefinite kernel function $k_\mathcal{T}:\mathcal{T}\times\mathcal{T} \rightarrow \mathbb{R}$. This corresponds to the assumption that each dimension of the vector-valued function $\wfunc$ is generated by an independent 
Gaussian process, and these Gaussian processes
share a common kernel function $k_\mathcal{T}$. 
Note that this decoupling is not an independence assumption on attributes; it is instead analogous to the assumption of an isotropic normal prior for model parameters that justifies the standard $\ell_2$-regularization.
We use $\bK_{\bT} \in \mathbb{R}^{n \times n}$ to denote the matrix
given by evaluations $k_{\mathcal{T}}(\bt_i,\bt_j)$ of the kernel function $k_{\mathcal{T}}$.
The process evaluates function $\wfunc$ for all $\bt_i$ to create parameter vectors $\bw_1=\wfunc(\bt_1), \dots,\bw_n=\wfunc(\bt_n)$.
The process then concludes by generating labels $y_i$ from an appropriate observation model,
\begin{equation}
\label{eq:observation_model}
y_i \sim p(y|\bx_i,\bw_i),
\end{equation}
for instance, a standard linear model with Gaussian noise for regression or a logistic function of the inner product of $\bw_i$ and $\bx_i$ for classification. 

The prediction problem is to infer the distribution of the label~$\newY$ for a new observation~$\newX$ with task variable~$\newT$. 
For notational convenience, we aggregate the training instances into matrix~$\bX\in\mathbb{R}^{n\times m}$ with row vectors~$\bx_1\trans,\dots,\bx_n\trans$, 
the task variables into matrix~$\bT\in\mathbb{R}^{n\times d}$
with row vectors~$\bt_1\trans,\dots,\bt_n\trans$, the parameter vectors associated with training observations into a matrix $\bW \in \mathbb{R}^{n \times m}$ with row vectors $\bw_1\trans,\dots,\bw_n\trans$, and the labels~$y_1,\dots,y_n$ into vector $\by \in \mathcal{Y}^n$.

In this model, the Gaussian process prior $p(\wfunc)$ over functions $\wfunc: \mathcal{T} \rightarrow \mathbb{R}^m$ couples 
parameter vectors $\wfunc(\bt)$ for different values $\bt$ of the task variable.
The hierarchical Bayesian model of multitask learning assumes a coupling of parameters 
based on a hierarchical Bayesian prior~\citep[\emph{e.g.},][]{gelman1995bayesian,finkel2009hierarchical}.
We will now show that the varying-coefficient model with isotropic GP prior subsumes hierarchical Bayesian multitask models by choice of an appropriate kernel 
function $\bkappa$ of the Gaussian process that defines $p(\wfunc)$.
Together with results on inference presented in Section~\ref{sec:inference}, this result shows how inference for hierarchical Bayesian multitask models 
can be carried out using a Gaussian process.

The following definition formalizes the hierarchical Bayesian multitask model.
\begin{definition}[Hierarchical Bayesian Multitask Model]
\label{def:hb}
Let $\mathcal{G} = (\mathcal{T},\bA)$ denote a tree structure over a set of tasks \mbox{$\mathcal{T} = \{1,\dots,k\}$} given by an adjacency matrix $\bA$,
with $1 \in \mathcal{T}$ the root node.
Let $\bsigma \in \mathbb{R}^k$ denote a vector with entries $\sigma_1,\dots,\sigma_k$.
The following process generates the distribution $p(\by|\bX,\bT;\mathcal{G},\bsigma)$ over labels $\by \in \mathcal{Y}^n$ given instances $\bX$, task variables $\bT$, the task hierarchy $\mathcal{G}$, and variances $\bsigma$: The process first samples 
parameter vectors $\wHB_{1},\dots,\wHB_{k} \in \mathbb{R}^m$ according to 
\begin{align}
&\wHB_{1} \sim \mathcal{N}(\wHB|\mathbf{0},\sigma_1^2 \eyeMatrix{m}) \label{eq:hb1}\\
&\wHB_{l} \sim \mathcal{N}(\wHB|\wHB_{pa(l)},\sigma_l^2 \eyeMatrix{m}) \hspace{1cm} 2 \leq l \leq k \label{eq:hb2}
\end{align}
where for $l \in \mathcal{T}$, $pa(l) \in \mathcal{T}$ is the unique node with $\bA_{pa(l),l} = 1$;
then, the process generates labels \mbox{$y_i \sim p(y|\bx_i,\wHB_{i})$}, where 
$p(y|\bx_i,\wHB_{i})$ is the same conditional distribution over labels given an instance 
and a parameter vector 
as
was chosen for the varying-coefficient model in Equation~\ref{eq:observation_model}.
This process defines the {\em hierarchical Bayesian multitask model}.
\end{definition}
The following proposition shows that the varying-coefficient model 
presented in Section~\ref{sec:model} subsumes the hierarchical Bayesian multitask model.
\begin{proposition}
\label{prop:hb_drift}
Let $\mathcal{G} = (\mathcal{T},\bA)$ denote a tree structure over a set of tasks $\mathcal{T} = \{1,\dots,k\}$ 
given by an adjacency matrix $\bA$.
Let $\bsigma \in \mathbb{R}^k$ be a vector with entries $\sigma_1,\dots,\sigma_k$.
Let $k_{\bA,\bsigma}: \mathcal{T} \times \mathcal{T} \rightarrow \mathbb{R}$ be given by
$k_{\bA,\bsigma}(t,t') = G_{t,t'}$, where 
$G_{i,j}$ denotes the entry at row $i$ and column $j$ of the matrix
\begin{align*}
\bG = (\eyeMatrix{k}-\bA)^{-1} \bS \left(\eyeMatrix{k}-\bA\trans\right)^{-1}, 
\end{align*}
and $\bS \in \mathbb{R}^{k \times k}$ 
denotes the diagonal matrix with entries $\sigma_1^2,\dots,\sigma_k^2$.
Let $\bkappa: \mathcal{T} \times \mathcal{T} \rightarrow \mathbb{R}^{m \times m}$ be given by 
\mbox{$\bkappa(t,t') = k_{\bA,\bsigma}(t,t') \eyeMatrix{m}$} and let
$p(\by|\bX,\bT;\bkappa) = \int p(\by|\bW,\bX)p(\bW|\bT;\bkappa)\dif \bW$ be the marginal distribution
over labels given instances and task variables defined by the varying-coefficient model.
Then it holds that $p(\by|\bX,\bT;\bkappa) = p(\by|\bX,\bT;\mathcal{G},\bsigma)$.
\end{proposition}
Proposition~\ref{prop:hb_drift} implies that performing Bayesian prediction in the varying-coefficient model with the specified kernel function is identical to performing Bayesian inference in the hierarchical Bayesian multitask model.
The proof is included in the appendix. 
In Proposition~\ref{prop:hb_drift}, entries $G_{t,t'}$ of $\bG$ represent a task similarity derived from the tree structure $\mathcal{G}$.
Instead of a tree structure over tasks, feature vectors describing individual tasks may also be given~\citep{bonilla2007kernel,Yan2009transfer}. In this case, $\bkappa(t,t')$
can be computed from the task features; the varying-coefficient model then subsumes existing approaches for multitask learning with task features
(see Section~\ref{sec:product}).

\section{Inference}
\label{sec:inference}

We now address the problem of inferring predictions $\newY$ for instances $\newX$, and task variables $\newT$. 
Section~\ref{sec:reg_problem} presents exact Bayesian solutions for regression; 
Section~\ref{sec:class_problem} discusses approximate Bayesian inference for classification. 
Section~\ref{sec:product} derives existing multitask models as special cases.

\subsection{Regression}
\label{sec:reg_problem}

This subsection studies 
linear regression models of the form $p(y | \bx, \bw) =  \mathcal{N}(y|\bx\trans\bw, \tau^2)$. 
Note that by substituting for the slightly heavier notation $p(y | \bx, \bw) =  \mathcal{N}(y|\Phi(\bx)\trans\bw, \tau^2)$, this treatment also covers finite-dimensional feature maps.
The predictive distribution for test instance $\newX$ with task variable $\newT$ is obtained by integrating over the possible parameter values~$\newW$ of the conditional distribution that has generated value $\newY$:
\begin{align}
p(\newY | \bX, \by, \bT, \newX, \newT) = \int p(\newY | \newX, \newW) p(\newW | \bX, \by, \bT, \newT)  \dif \newW,\label{eq:predictive_distribution}
\end{align}
where the posterior over $\newW$ is obtained by integrating over the joint parameter values $\bW$ that have generated the labels $\by$ for instances $\bX$ and task variables $\bT$:
\begin{align}
p(\newW | \bX, \by, \bT, \newT) = \int p(\newW| \bW, \bT, \newT)p(\bW | \bX, \by, \bT)  \dif \bW. \label{eq:integrating_W}
\end{align}
Posterior distribution $p(\bW | \bX, \by, \bT)$ in Equation \ref{eq:integrating_W} depends on the likelihood function---the linear model---and the GP prior $p(\wfunc)$. The extrapolated posterior $p(\newW| \bW, \bT, \newT)$ for test instance $\newX$ with task variable $\newT$ depends on the Gaussian process.
The following theorem states how the predictive distribution given by Equation~\ref{eq:predictive_distribution} can be computed.

\begin{theorem}[Bayesian Predictive Distribution]
\label{th:predDist}
Let $\mathcal{Y} = \mathbb{R}$, $p(y|\bx,\bw) = \mathcal{N}(y|\bx\trans \bw, \tau^2)$,
and let the kernel matrix~~$\bK_{\bT}$ be positive definite.
Let~$\bK \in \mathbb{R}^{n\times n}$ be a matrix with components \mbox{$k_{ij}=\bx_i\trans\bx_jk_\mathcal{T}(\bt_i,\bt_j)$} 
and $\bk \in \mathbb{R}^n$ be a vector with components \mbox{$k_{i}=\bx_i\trans\newX k_\mathcal{T}(\bt_i,\newT)$}.
Then, the predictive distribution for the 
varying-coefficient model defined in Section~\ref{sec:model} is given~by
\begin{align}
	p(\newY|\bX,\by,\bT, \newX,\newT) &= \mathcal{N}(\newY|\mu,\sigma^2+\tau^2)\label{eq:ypost}
\end{align}
with
\begin{align*}
  \mu &= \bk\trans(\bK+\tau^2 \eyeMatrix{n})^{-1}\by,\\
	\sigma^2 &= \newX\trans \newX k_\mathcal{T}(\newT,\newT) - \bk\trans(\bK+\tau^2\eyeMatrix{n})^{-1}\bk. 
\end{align*}
\end{theorem}
Before we prove Theorem~\ref{th:predDist}, we highlight three observations about this result.
First, the distribution $p(\newY|\bX,\by,\bT, \newX,\newT)$
has a surprisingly simple form. It is identical to the predictive distribution of a standard Gaussian process 
that uses  concatenated vectors $(\bx_1,\bt_1),\dots,(\bx_n,\bt_n) \in \mathcal{X}\times \mathcal{T}$ as training instances, 
labels $y_1,\dots,y_n$, and the product kernel function $k((\bx_i,\bt_i),(\bx_j,\bt_j)) = \bx_i\trans \bx_j k_\mathcal{T}(\bt_i,\bt_j)$.

Secondly, instances $\bx_1,\dots,\bx_n,\newX \in \mathcal{X}$ only
enter Equation~\ref{eq:ypost} in the form of inner products. The model can therefore directly be kernelized 
by defining the kernel matrix as $\bK_{ij}=k_{\mathcal{X}}(\bx_i,\bx_j) k_\mathcal{T}(\bt_i,\bt_j)$ with kernel function $k_{\mathcal{X}}(\bx_i,\bx_j) = \Phi(\bx_i)\trans \Phi(\bx_j)$ where
$\Phi$ 
maps to a 
reproducing kernel Hilbert space. When the feature space is finite, then $\wfunc$ maps the $\bt_i$ to a finite-dimensional $\bw_i$ and Theorem~\ref{th:predDist}  implies a Bayesian predictive distribution derived from the generative process that Section~\ref{sec:model} specifies. When the reproducing kernel Hilbert space does not have a finite dimension, Section~\ref{sec:model} does no longer specify a corresponding proper generative process because $p(\bw_1,\dots,\bw_n|\bT)$ would otherwise become infinite-dimensionally normally distributed. However, given the finite sample $\bX$ and $\bT$, a Mercer map~\citep[see,~\eg][Section 2.2.4]{scholkopf2002learning} constitutes a finite-dimensional space $\mathbb{R}^n$ for which Section~\ref{sec:model} again characterizes a corresponding generative process.

Thirdly and finally, Theorem~\ref{th:predDist} shows how Bayesian inference in varying-coefficient models with isotropic priors can be implemented much more efficiently than in general varying-coefficient models.
Bayesian inference in varying-coefficient models in the parameter space generally involves matrices of size $nm \times nm$ because it needs to take the overall covariance structure into account; the algorithm of Gelfand et al.
infers the covariance matrix under an inverse Wishart prior using a sliced Gibbs sampler over parameter values~\cite{gelfand2003spatial}.
This makes inference impractical for large-scale problems. 
Theorem~\ref{th:predDist} shows that under the isotropy assumption, the latent parameter vectors $\bw_1,\dots,\bw_n$ can be integrated out, which results in a GP formulation in which the covariance structure over parameter vectors resolves to an $n \times n$ product-kernel matrix.

\begin{proof}\textbf{of Theorem~\ref{th:predDist}}.
Let $w_{ir}$ and $w_{\star r}$ denote the $r$-th elements of vectors $\bw_i$ and $\newW$, and let 
$x_{ir}$ and $x_{\star r}$ denote the $r$-th elements of vectors $\bx_i$ and $\newX$. 
Let $\bz_{\star} = (z_1,\ldots,z_n,z_{\star})\trans \in \mathbb{R}^{n+1}$ with $z_i = \bx_i\trans\bw_i$ and $z_{\star} = \newX\trans\newW$.
Because $\bw_1,\ldots,\bw_n,\newW$ are evaluations of the function $\wfunc$ drawn from a Gaussian process (Equation~\ref{eq:gp}), 
they are jointly Gaussian distributed and thus $z_1,\ldots,z_n,z_{\star}$ are also jointly Gaussian~\citep[\eg][Chapter 10.2.5]{murphy2012machine}.
Because $\wfunc$ is drawn from a zero-mean process, it holds that
$\mathbb{E}[z_i] = \mathbb{E}[\sum_{r=1}^m{x_{ir} w_{ir}}] = \sum_{r=1}^m{x_{ir}\mathbb{E}[w_{ir}]} = 0$ as well as $\mathbb{E}[z_{\star}] = 0$ and therefore
\begin{equation*}
p(\bz_{\star}|\bX,\bT,\newX,\newT) = \mathcal{N}(\bz_{\star}|\mathbf{0},\mathbf{C})
\end{equation*}
where $\bC \in \mathbb{R}^{(n+1) \times (n+1)}$ denotes the covariance matrix. 
For the covariances $\mathbb{E}[z_i z_j]$ it holds that
\begin{align}
\mathbb{E} \left[z_i z_j\right] &= \mathbb{E} \left[ \bx_i\trans\bw_i  \bx_j\trans\bw_j \right] \notag \\
&= \mathbb{E} \left[ \left(\sum_{s=1}^m x_{is}w_{is} \right) \left(\sum_{r=1}^m x_{jr}w_{jr}\right) \right] \notag  \\
&= \sum_{s=1}^m \sum_{r=1}^m x_{is} x_{jr} \mathbb{E} \left[ w_{is} w_{jr} \right] \notag \\
&= \sum_{s=1}^m x_{is} x_{js} \mathbb{E} \left[ w_{is} w_{js} \right] \label{eq:exploit_isotropy1}\\
&= \bx_i\trans \bx_j k_{\mathcal{T}}(\bt_i,\bt_j). \label{eq:exploit_isotropy2}
\end{align}
In Equations~\ref{eq:exploit_isotropy1} and~\ref{eq:exploit_isotropy2} we exploit the isotropy of the Gaussian process prior:
the covariance $\mathbb{E}[w_{is} w_{jr}]$ is
the element in row $s$ and column $r$ of the matrix \mbox{$\bkappa(\bt_i,\bt_j) \in\mathbb{R}^{m \times m}$} obtained by evaluating the kernel function 
\mbox{$\bkappa:\mathcal{T}\times\mathcal{T} \rightarrow \mathbb{R}^{m \times m}$ at $(\bt_i,\bt_j)$};  
the isotropy assumption $\bkappa(\bt,\bt') = k_{\mathcal{T}}(\bt,\bt') \eyeMatrix{m}$ means that this matrix is 
diagonal with $\mathbb{E}[w_{is} w_{jr}] = 0$ for $s \neq r$ and $\mathbb{E}[w_{is} w_{js}] = k_{\mathcal{T}}(\bt_i,\bt_j)$  (see Section~\ref{sec:model}).
We analogously derive 
\begin{align}
\mathbb{E}[z_i z_{\star}] =  \bx_i \trans \newX k_{\mathcal{T}}(\bt_i,\newT), \label{eq:product_i_star}\\
\mathbb{E}[z_{\star} z_{\star}] =  \newX \trans \newX k_{\mathcal{T}}(\newT,\newT). \label{eq:product_star_star}
\end{align}
Equations~\ref{eq:exploit_isotropy2},~\ref{eq:product_i_star} and~\ref{eq:product_star_star} define the covariance matrix $\bC$, yielding 
\begin{equation*}
p(\bz_{\star}|\bX,\bT,\newX,\newT) = \mathcal{N}\left(\bz_{\star}|\mathbf{0},\left( 
\begin{array}{cc}
\bK & \bk \\
\bk\trans & k_{\star}
\end{array} \right) \right)
\end{equation*}
where $k_{\star} = \newX\trans\newX k_{\mathcal{T}}(\newT,\newT)$.
For $\by_{\star} = (y_1,\ldots,y_n,\newY)$ it now follows that 
\begin{align}
p(\by_{\star}|&\bX,\bT,\newX,\newT) = 
\mathcal{N}\left(\by_{\star}|\mathbf{0},\left( 
\begin{array}{cc}
\bK+\tau^2\eyeMatrix{n} & \bk\\
\bk\trans & k_{\star}+\tau^2
\end{array} \right) \right).\label{eq:joint_distribution_all_ys}
\end{align}
The claim now follows by applying standard Gaussian identities to compute the conditional distribution $p(\newY|\bX,\by,\bT,\newX,\newT)$ from Equation~\ref{eq:joint_distribution_all_ys}.
\end{proof}

\subsection{Classification}
\label{sec:class_problem}

The result given by Theorem~\ref{th:predDist} can be extended to classification settings with $\mathcal{Y} = \{0,1\}$
by using non-Gaussian likelihoods $p(y|z)$ that generate labels $y \in \mathcal{Y}$ given outputs $z \in \mathbb{R}$ of the linear model.
\begin{theorem}[Bayesian predictive distribution for non-Gaussian likelihoods]
\label{th:predDistClass}
Let $\mathcal{Y} = \{0,1\}$. Let $p(y_i|\bx_i,\bw_i)$ be given by a generalized linear model, defined 
by $z_i~\sim~\mathcal{N}(z|\bw_i\trans \bx_i,\tau^2)$ and $y_i \sim p(y|z_i)$. Let 
$p(\newY|\newX,\newW)$ be given by \mbox{$\newZ \sim \mathcal{N}(z|\newW\trans \newX,\tau^2)$} and
$\newY \sim p(y|\newZ)$. Let furthermore $\bz = (z_1,\ldots,z_n)\trans \in \mathbb{R}^n$.

Let the kernel matrix~$\bK_{\bT}$ be positive definite, and let~$\bK \in \mathbb{R}^{n\times n}$ be a matrix with components \mbox{$k_{ij}=\bx_i\trans\bx_jk_\mathcal{T}(\bt_i,\bt_j)$} and~$\bk \in \mathbb{R}^n$ a vector with components \mbox{$k_{i}=\bx_i\trans\newX k_\mathcal{T}(\bt_i,\newT)$}.
Then, the predictive distribution for the 
GP model defined in Section~\ref{sec:model} is given~by
\begin{align}
p(\newY|\bX,\by,\bT,\newX,\newT) \propto
 \iint p(\newY|\newZ) \mathcal{N}(\newZ|\mu_{\bz},\sigma_{\bz}^2) p(\by|\bz) \mathcal{N}(\bz|\mathbf{0},
\bK   + \tau^2 \eyeMatrix{n}) \dif \bz \dif \newZ \label{eq:predDistClass}
\end{align}
with
\begin{align*}
	\mu_{\bz} &= \bk\trans(\bK+\tau^2 \eyeMatrix{n})^{-1}\bz,\\
	\sigma_{\bz}^2 &= \newX\trans \newX k_\mathcal{T}(\newT,\newT) - \bk\trans(\bK+\tau^2\eyeMatrix{n})^{-1}\bk + \tau^2.
\end{align*}
\end{theorem}
A straightforward calculation shows that Equation~\ref{eq:predDistClass} is identical to
the predictive distribution of a standard Gaussian process that uses concatenated 
vectors $(\bx_1,\bt_1),\dots,(\bx_n,\bt_n) \in \mathcal{X}\times \mathcal{T}$ as training instances, 
labels $y_1,\dots,y_n$, the product kernel $k((\bx_i,\bt_i),(\bx_j,\bt_j)) = \bx_i\trans \bx_j k_\mathcal{T}(\bt_i,\bt_j)$,
and likelihood function $p(y|z)$.
For non-Gaussian likelihoods, exact inference in Gaussian processes is generally intractable, but approximate inference methods based on, \emph{e.g.}, Laplace approximation, variational inference or expectation propagation are available.

\sloppy
\begin{proof}\textbf{of Theorem~\ref{th:predDistClass}}.
Rewriting $p(\newY|\bX,\by,\bT,\newX,\newT)$ in terms of a marginalization over the variables $\bz$ and $\newZ$ leads to:
\begin{align*}
p(\newY|\bX,\by,\bT,\newX,\newT)
& =\int{p(\newY|\newZ)p(\newZ|\bX,\by,\bT,\newX,\newT) \dif \newZ} \\
& =\iint{p(\newY|\newZ)p(\newZ|\bX,\bz,\bT,\newX,\newT)p(\bz|\bX,\by,\bT)\dif\bz\dif \newZ}\\
& \propto  \iint{ p(\newY|\newZ)p(\newZ|\bX,\bz,\bT,\newX,\newT)p(\by|\bz)p(\bz|\bX,\bT)\dif\bz\dif \newZ}.
\end{align*} 
The proof now quickly follows from Theorem~\ref{th:predDist} and derivations in the proof of Theorem~\ref{th:predDist}:
Equation~\ref{eq:ypost} implies $p(\newZ|\bX,\bz,\bT,\newX,\newT) =  \mathcal{N}(\newZ|\mu_{\bz},\sigma_{\bz}^2)$,
Equation~\ref{eq:joint_distribution_all_ys} implies \mbox{$p(\bz|\bX,\bT) = \mathcal{N}(\bz|\mathbf{0},\bK + \tau^2 \eyeMatrix{n})$}.
\end{proof}

\fussy

\subsection{Product Kernels in Transfer Learning }\label{sec:product}

Sections~\ref{sec:reg_problem} and~\ref{sec:class_problem} have shown that inference in the varying-coefficient model is equivalent
to inference in standard Gaussian processes with products of task kernels and instance kernels. 
Similar product kernels are used in several existing transfer learning models. 
Our results identify the generative assumptions that underlie these models by showing that the product kernels
which they employ can be derived from the assumption of a varying-coefficient model with isotropic GP prior and an appropriate kernel function.

\citet{bonilla2007kernel} study a setting in which there is a discrete set of $k$ tasks, which are described 
by task-specific attribute vectors $\bt_1,\dots,\bt_k$. They study a Gaussian process model based on concatenated feature
vectors $(\bx,\bt)$ and
a product kernel $k((\bx,\bt),(\bx',\bt')) = k_\mathcal{X}(\bx,\bx')k_{\mathcal{T}}(\bt,\bt')$, where 
$k_\mathcal{X}(\bx,\bx')$ reflects instance similarity and $k_{\mathcal{T}}(\bt,\bt')$ reflects task similarity. Theorems~\ref{th:predDist} and~\ref{th:predDistClass} identify
the generative assumptions underlying this model: a varying-coefficient model with isotropic Gaussian process prior and kernel $k_{\mathcal{T}}$ generates task-specific parameter vectors in a reproducing Hilbert space of the instance kernel $k_{\mathcal{X}}$; a linear
model in that Hilbert space generates the observed labels.
 
\citet{evgeniou2005learning} and~\citet{alvarez2011kernels} study multitask-learning problems in which task similarities are given in terms of a task graph. 
Their method uses the product of an instance kernel and the graph-Laplacian kernel of the task graph. We will now show that, when the task graph is a tree, that kernel emerges from Proposition \ref{prop:hb_drift}. This signifies that, when the task graph is a tree, the graph regularization method of~\citet{evgeniou2005learning} is the dual formulation of hierarchical Bayesian multitask learning, and therefore Bayesian inference for hierarchical Bayesian models can be carried out efficiently using a standard Gaussian process with a graph-Laplacian kernel.
\begin{definition}[Graph-Laplacian Multitask Kernel]
\label{def:graph_laplacian_kernel}
Let $\mathcal{G} = (\mathcal{T},\bM)$ denote a weighted undirected graph structure over a set of tasks $\mathcal{T} = \{1,\dots,k\}$ given by a symmetric adjacency 
matrix \mbox{$\bM \in \mathbb{R}^{k \times k}$}, where
$\bM_{i,j}$ defines the positive weight of the edge between tasks $i$ and $j$ or $\bM_{i,j}=0$ if no such edge exists.
Let $\bD$ denote the weighted degree matrix of the graph, and $\bL = \bD + \bR - \bM$ the graph Laplacian, 
where a diagonal matrix $\bR$ that acts as a regularizer has been added to the degree matrix~\citep{alvarez2011kernels}.
The kernel function $k_{\bM,\bR}: \left(\mathcal{X}\times \mathcal{T}\right) \times \left(\mathcal{X}\times \mathcal{T}\right) \rightarrow \mathbb{R}$ 
given by 
\[ k_{\bM,\bR}((\bx,\bt),(\bx',\bt')) = \bL^{\dagger}_{\bt,\bt'} \bx \trans \bx',\]
 where $\bL^{\dagger}$ is the pseudoinverse of~~$\bL$, will be referred
to as the \emph{graph-Laplacian multitask kernel}.
\end{definition}

The following proposition states that the graph-Laplacian multitask kernel is equal to the kernel that emerges in the dual formulation of hierarchical Bayesian multitask learning (Definition~\ref{def:hb}).
\begin{proposition}
\label{prop:hb_laplace}
Let $\mathcal{G}=(\mathcal{T},\bA)$ denote a directed tree structure given by an adjacency matrix $\bA$. 
Let $\bsigma \in \mathbb{R}^k$ be a vector with entries $\sigma_1,\dots,\sigma_k$.
Let $\bB \in \mathbb{R}^{k \times k}$ denote the diagonal matrix with entries $0,\sigma_2^{-2},\dots,\sigma_k^{-2}$,
let $\bR \in \mathbb{R}^{k \times k}$ denote the diagonal matrix with entries $\sigma_1^{-2},0,\dots,0$,
let $\bM = \bB \bA + (\bB\bA)\trans$, and let $k_{\bA,\bsigma}(\bt,\bt')$ be defined as in Proposition~\ref{prop:hb_drift}.
Then 
\[ k_{\bM,\bR}((\bx,\bt),(\bx',\bt')) = k_{\bA,\bsigma}(\bt,\bt')\bx\trans\bx'.\]
\end{proposition}
Note that in Proposition~\ref{prop:hb_laplace}, $\bB \bA$ is an adjacency matrix in which an edge from node $i$ to node $j$ is weighted 
by the respective precision $\sigma_j^{-2}$ of the conditional distribution (Equation~\ref{eq:hb2});
adding the transpose yields a symmetric matrix $\bM$ of task relationship weights.
The precision $\sigma_1^{-2}$ of the root node prior is subsumed in the regularizer $\bR$.
The proof is included in the appendix.

\section{Empirical Study}
\label{sec:exp}

In this section, we study the efficiency and accuracy of different varying-coefficient models and baselines for geospatial and temporal regression and classification problems. 
We focus on the problems of predicting real estate prices and monthly housing rents.

For real estate price prediction, we acquire records of real-estate sales in New York City  
for sales dating from January 2003 to December 2009 in June 2013 through the NYC Open Data initiative\footnote{\url{https://nycopendata.socrata.com/}.}
. Input variables include the floor space, plot area, property class (such as family home, residential condominium, office, or store), date of construction of the building, and the number of residential and commercial units in the building. After binarization of multi-valued attributes there are 94 numeric attributes in the data set.
For regression, the sales price serves as target variable $y$; we also study a classification problem in which $y$ is a binary indicator that distinguishes between transactions with a price above the median of 450,000 dollars from transactions below it.
Date and address for every sale are available; we transform addresses into geographical latitude and longitude using an inverse geocoding service based on OpenStreetMap data.
We encode the sales date and geographical latitude and longitude of the property as task variable $\bt \in \mathbb{R}^3$. 

Price and attributes in sales records vary widely; for instance, prices range from one dollar to four billion dollars, and the floor space from one square foot to fourteen million square feet. A substantial number of records contain either errors or document transactions in which the valuations do not reflect the actual market values: for instance, Manhattan condominiums that sold for one dollar, and one-square-foot lots that sold for massive prices. 
In order to filter most off-market transactions by means of a simple policy, we only include records of sales within a price range of 100,000 to 1,000,000 dollars, a property area range of 500 to 5,000 square feet, and a land area range of 500 to 10,000 square feet. Approximately 80\% of all records fall into these brackets. Additionally, we remove all records with missing values. After preprocessing, the data set contains 231,708 sales records.
We divide the records, which span dates from January 2003 to December 2009, into 25 consecutive blocks. 
Models are trained on a set of $n$ instances sampled randomly from a window of five blocks of historical data and evaluated on the subsequent block; results are averaged over all blocks.

For rent prediction, we acquire records on the monthly rent paid for privately rented apartments and houses in the states of California and New York from the 2013 American Community Survey's ASC public use microdata sample files%
\footnote{\url{http://factfinder.census.gov/faces/affhelp/jsf/pages/metadata.xhtml?lang=en&type=document&id=document.en.ACS_pums_csv_2013\#main_content}.}. 
Input variables include the number of rooms, number of bedrooms, the duration for which the contract has been running, the construction year of the building, the type of building (mobile home, trailer, or boat; attached or detached family house; apartment building), and variables that describe technical facilities (\eg variables related to internet access, type of plumbing, and type of heating). After binarization of multi-valued attributes there are 24 numerical attributes in the data. 
We study a regression problem in which the target variable $y$ is the monthly rent, and a classification problem in which $y$ is a binary indicator that distinguishes contracts with a monthly rent above the median of 1,200 dollars from those with a rent below the median. 
For each record, the geographical location is available in the form of a public use microdata area (PUMA) code\footnote{\url{https://www.census.gov/geo/reference/puma.html}.}. We translate PUMA codes to geographical latitude and longitude by associating each record with the longitude-latitude-centroid of the corresponding public use microdata area; 
these geographical latitudes and longitudes constitute the task variable $\bt \in \mathbb{R}^2$.
 We remove all records with missing values. The preprocessed data sets contain 36,785 records (state of California) and 17,944 records (state of New York).
Models are evaluated using 20-fold cross validation; in each fold, a random subset of $n$ training instances is sampled randomly from the respective training fold.

We study the varying-coefficient model with isotropic GP prior introduced in Section~\ref{sec:model} with a Mat\'{e}rn kernel $k_{\mathcal{T}}(\bt,\bt')$.
Predictions are obtained from Theorem~\ref{th:predDist}, using either a linear or also a Mat\'{e}rn 
kernel function $k_{\mathcal{X}}(\bx,\bx')$ 
(denoted  by \ourModelLinear \ and \ourModelRBF, respectively).
We compare with the varying-coefficient model with nonisotropic GP prior by~\citet{gelfand2003spatial}, in which the covariances are inferred from data (denoted by \textsl{Gelfand}).
Furthermore, we compare with the kernel-local smoothing varying-coefficient model of~\citet{fan2008statistical} that infers point estimates of model parameters.
We study this model using a linear feature map (\baselineFanLinear) and a nonlinear feature map constructed from a Mat\'{e}rn  kernel (\baselineFanRBF).
\citet{fan2008statistical} do not regularize parameter estimates in their original model, we added an $\ell_2$-regularizer as this improved 
predictive performance. 

\begin{figure}[t]
	\begin{center}
	 \vspace{2mm}
	 \centering
		\includegraphics[width=0.45\linewidth]{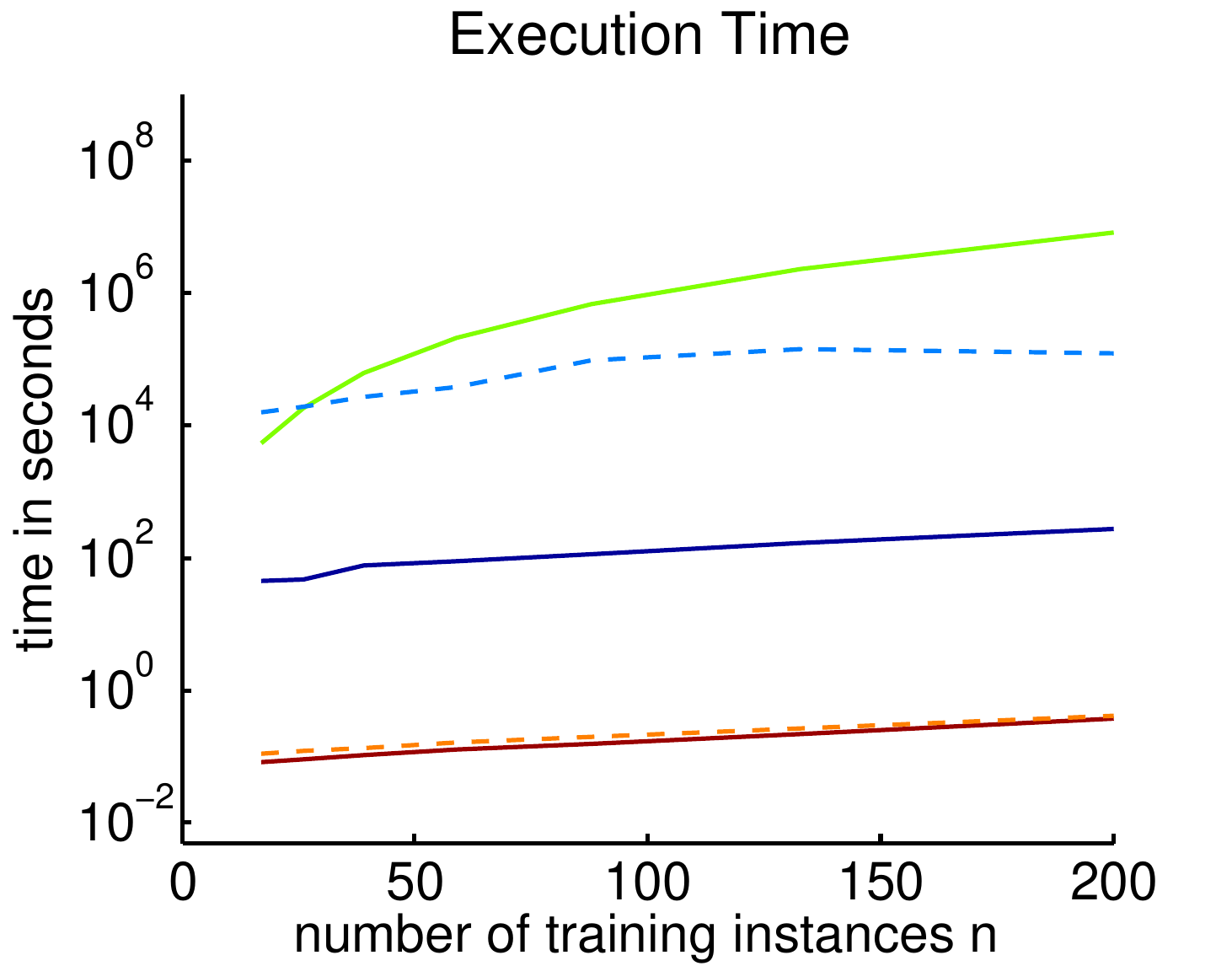}
		\hspace{5mm}
		\begin{minipage}{0.23\linewidth}
		\vspace{-4.5cm}
		\fbox{\includegraphics[width=\linewidth, viewport=30 80 138 132]{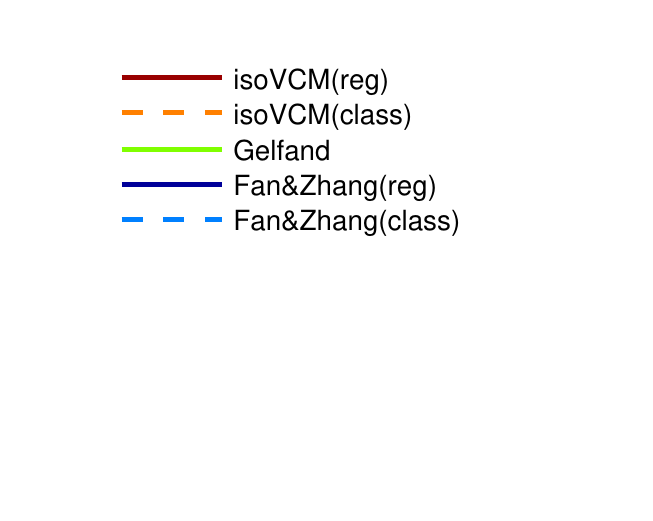}}
	 \vspace{0mm}
	 \end{minipage}
	 		\caption{
				Execution time of \ourModel and reference methods 
				over training set size $n$.
			}
		\label{fig:graphical_model_runtime}
	\end{center}
	\vskip -0.1in
\end{figure} 
We finally compare against an \emph{iid} baseline that assumes that $p(y|\bx)$ is constant in $\bt$, implemented by a standard Gaussian process with a linear (\baselineGPNaiveLinear) or Mat\'{e}rn (\baselineGPNaiveRBF) kernel, and with a standard Gaussian process that simply concatenates instance and task attribute vectors into vectors $(\bx, \bt)$ (denoted \baselineGPTimeGeoLinear and \baselineGPTimeGeoRBF).

For classification, we use logistic likelihood functions in our model (Theorem~\ref{th:predDistClass}),
and also in the GP baselines and the kernel-local smoothing varying-coefficient model of~\citet{fan2008statistical}.
All kernel parameters, as well as the observation noise parameter $\tau$ of Theorem~\ref{th:predDist} and the 
observation noise parameters of the standard GP models are tuned according to marginal likelihood on the training data. 
The regularization parameter of the kernel-local smoothing varying-coefficient model and its kernel parameter $h$ (see~\citealp{fan2008statistical})
are tuned on the training data by cross-validation.
The \ourModel model and all GP baselines are implemented based on the GPML Gaussian process toolbox~\citep{rasmussen2010gaussian}. 
Inference is carried out using the FITC approximation based on a low-rank approximation to the exact covariance matrix with 1,000 randomly sampled inducing 
points~\citep{snelson2006sparse}, and using Laplace approximation for classification.

First, we compare the execution time of the GP inference that results from Theorem \ref{th:predDist} with the execution time of the primal inference procedure of~\citet{gelfand2003spatial}
and the execution time of the kernel-local smoothing varying-coefficient model of~\citet{fan2008statistical}.
Figure~\ref{fig:graphical_model_runtime} shows the execution time for model training and prediction on one block of test instances in the real estate price prediction task as a function of the training set size $n$
(CPU core seconds, Intel Xeon 5520, 2.26 GHz).
For the model of Gelfand et al., the most expensive step during inference is computation of the inverse of a Cholesky decomposition of an $nm \times nm$ matrix, which 
needs to be performed within each Gibbs sampling iteration. Figure~\ref{fig:graphical_model_runtime} shows the execution time of 5,000 iterations of this step (3,000 burn-in and 2,000 sampling iterations, according to~\citealp{gelfand2003spatial}), yielding a lower bound on the overall execution time. 
An experimental run with Bayesian inference for nonisotropic GP priors requires 230 CPU core days even for 100 training instances; 
as matrix inversion scales nearly cubically in $n$, it is 
impractical for this application.
We therefore exclude this method from the remaining experiments. 
By contrast, full Bayesian inference in our GP model takes less than a second. 
The execution time of the kernel-local smoothing varying-coefficient model by~\citet{fan2008statistical} 
substantially differs for the regression and classification task. 
In this model, separate point estimates of model parameters have to be inferred for each test instance, for which a separate optimization problem needs to be solved. 
For regression, efficient closed-form solutions for parameter estimates are available, while for classification more expensive numerical 
optimization is required~\citep{fan2008statistical}.

\begin{figure*}[t]
	\begin{center}
	\vspace{2mm}
	  \includegraphics[width=0.49\linewidth]{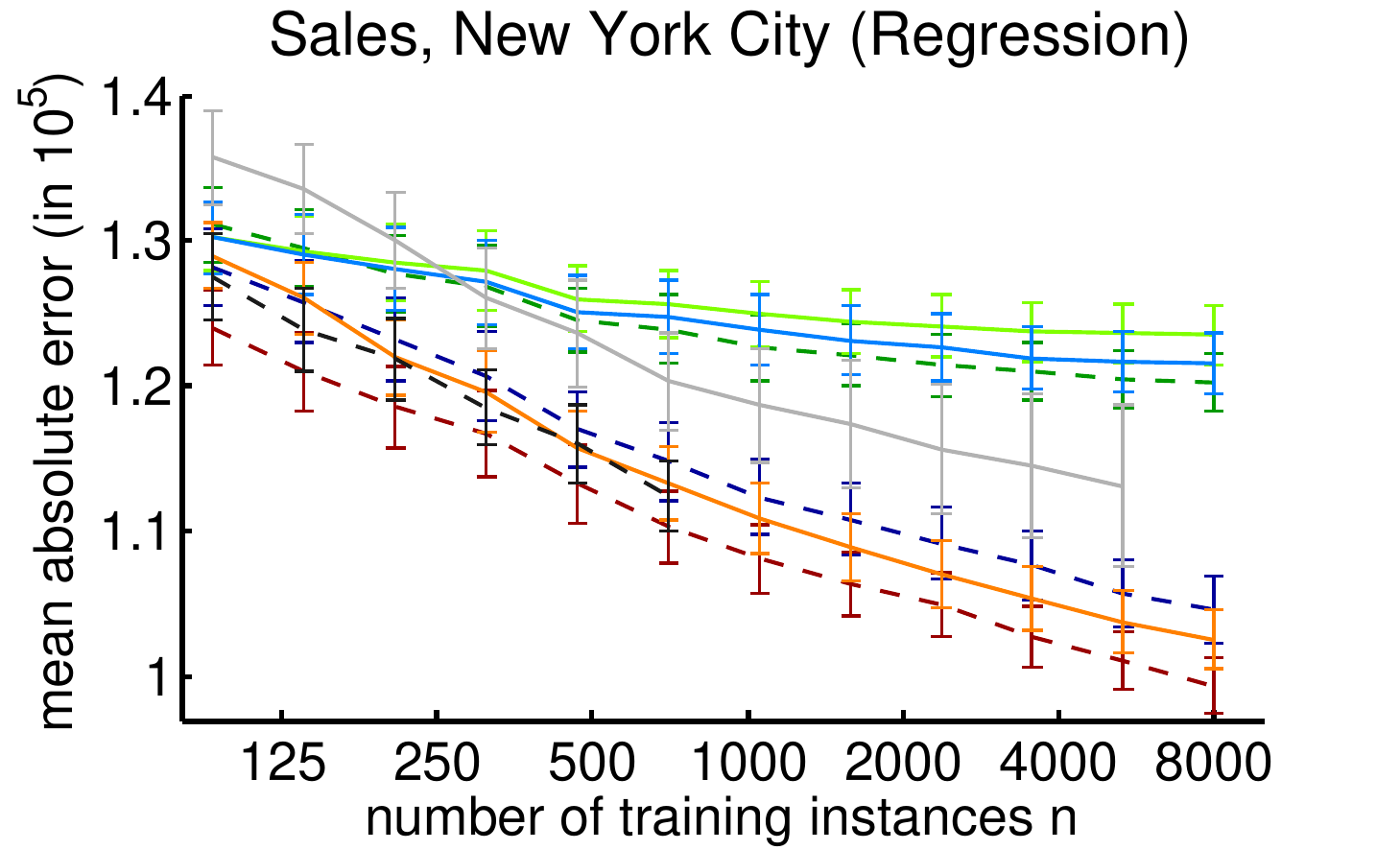}
	  \hspace{1mm}
		\includegraphics[width=0.49\linewidth]{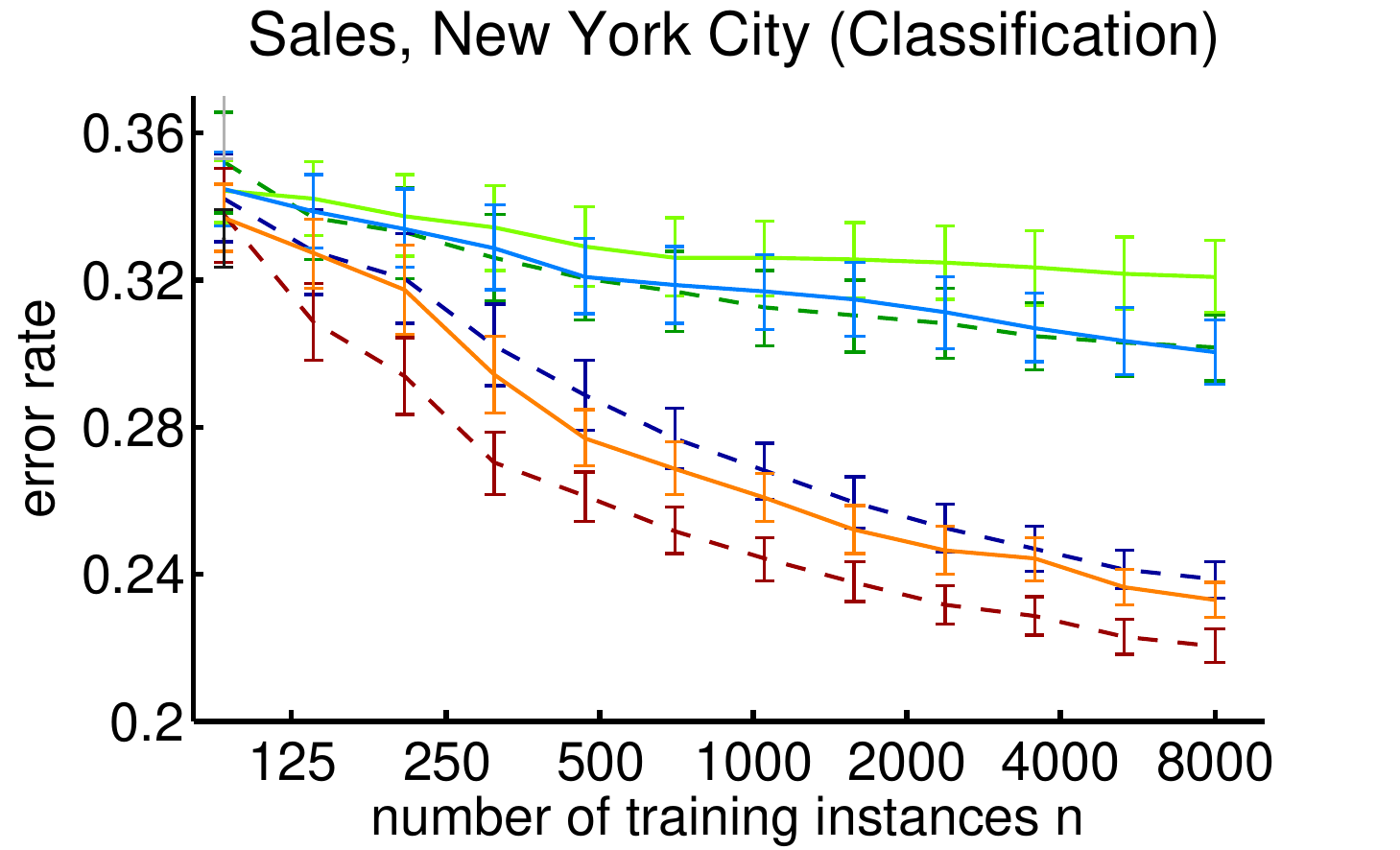}\\
		\vspace{5mm}
		\centering
		\fbox{\includegraphics[width=0.95\linewidth, viewport=0 52 560 72]{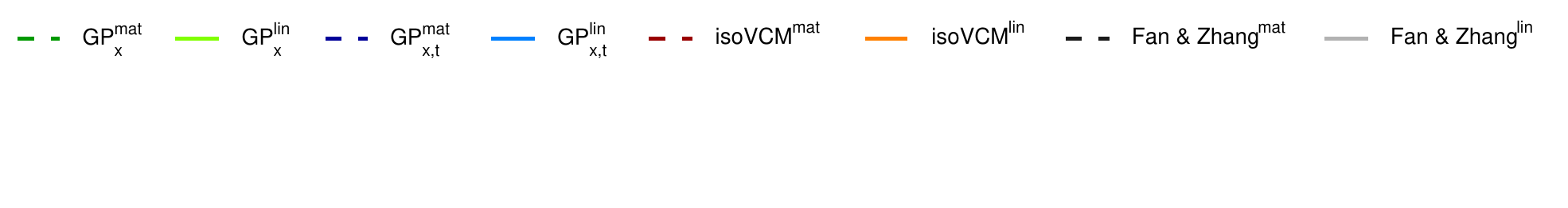}}
			\caption{
	Mean absolute error for predicting real estate prices in New York City (left) and mean zero-one loss for classifying real estate transactions (right) over training set size $n$.	Error bars indicate the standard error.
			}
		\label{fig:main_results}
	\end{center}
\end{figure*} 

In all subsequent experiments, each method is given 30 CPU core days 
of execution time; experiments are run sequentially for increasing number $n$ of training instances and 
results are reported for values of $n$ for which the cumulative execution time is below this limit.

\begin{figure*}[t]
	\begin{center}
	\vspace{1mm}
	  \includegraphics[width=0.49\linewidth]{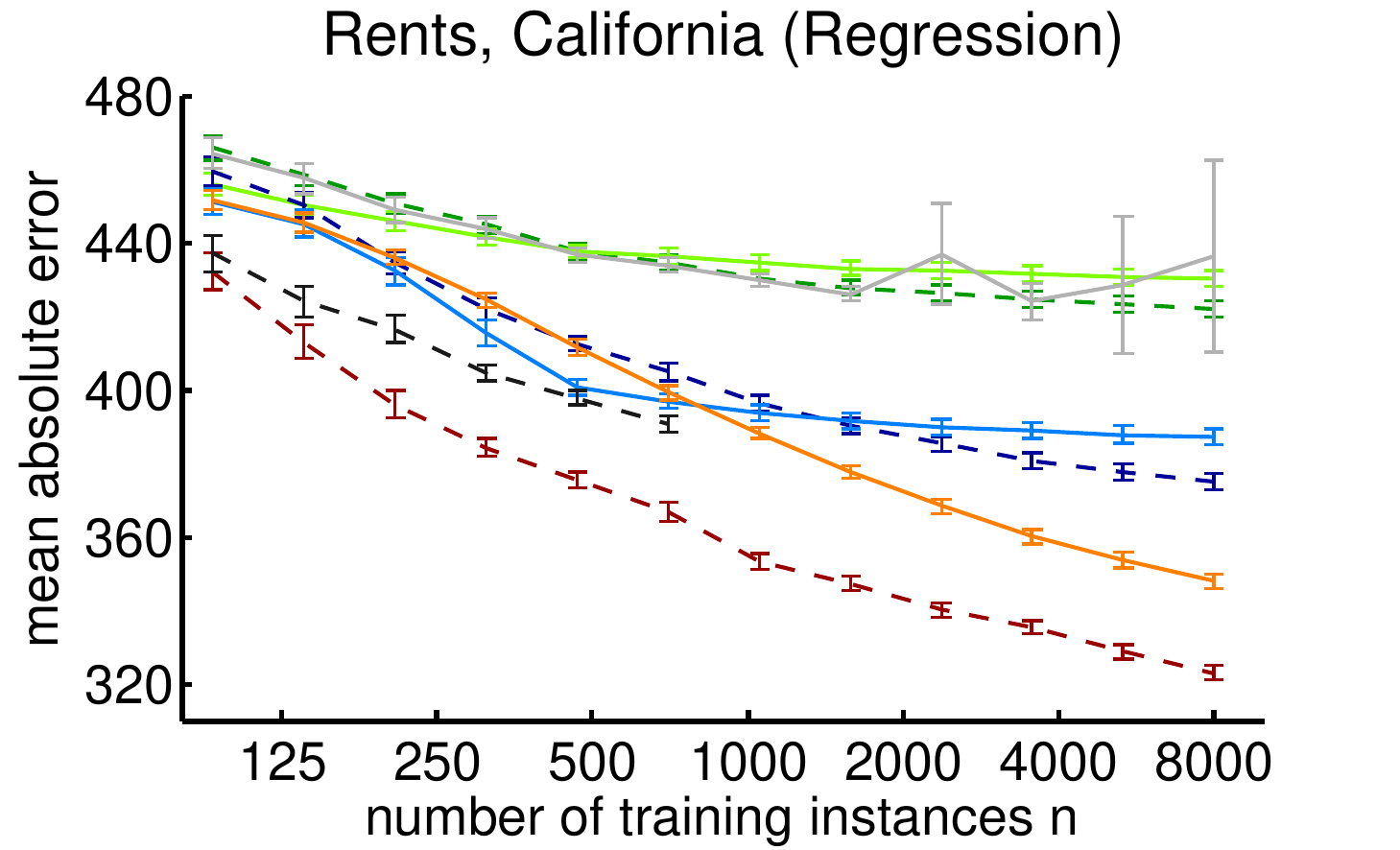}
		\hspace{1mm}
		\includegraphics[width=0.49\linewidth]{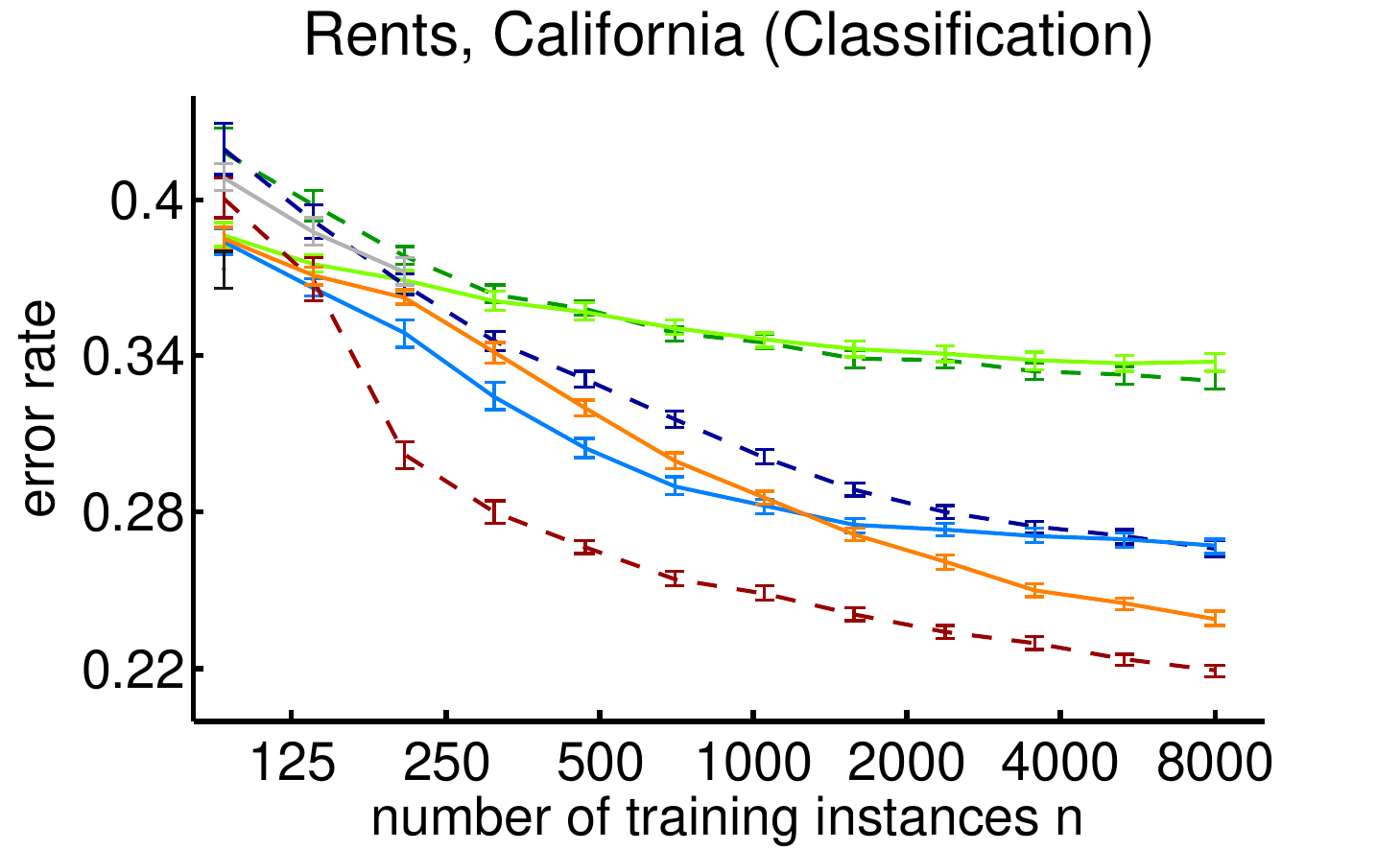}\\
		\vspace{5mm}
		\centering
		\fbox{\includegraphics[width=0.95\linewidth, viewport=0 52 560 72]{figures//LegendHorizontal}}\\
		\vspace{9mm}
	  \includegraphics[width=0.49\linewidth]{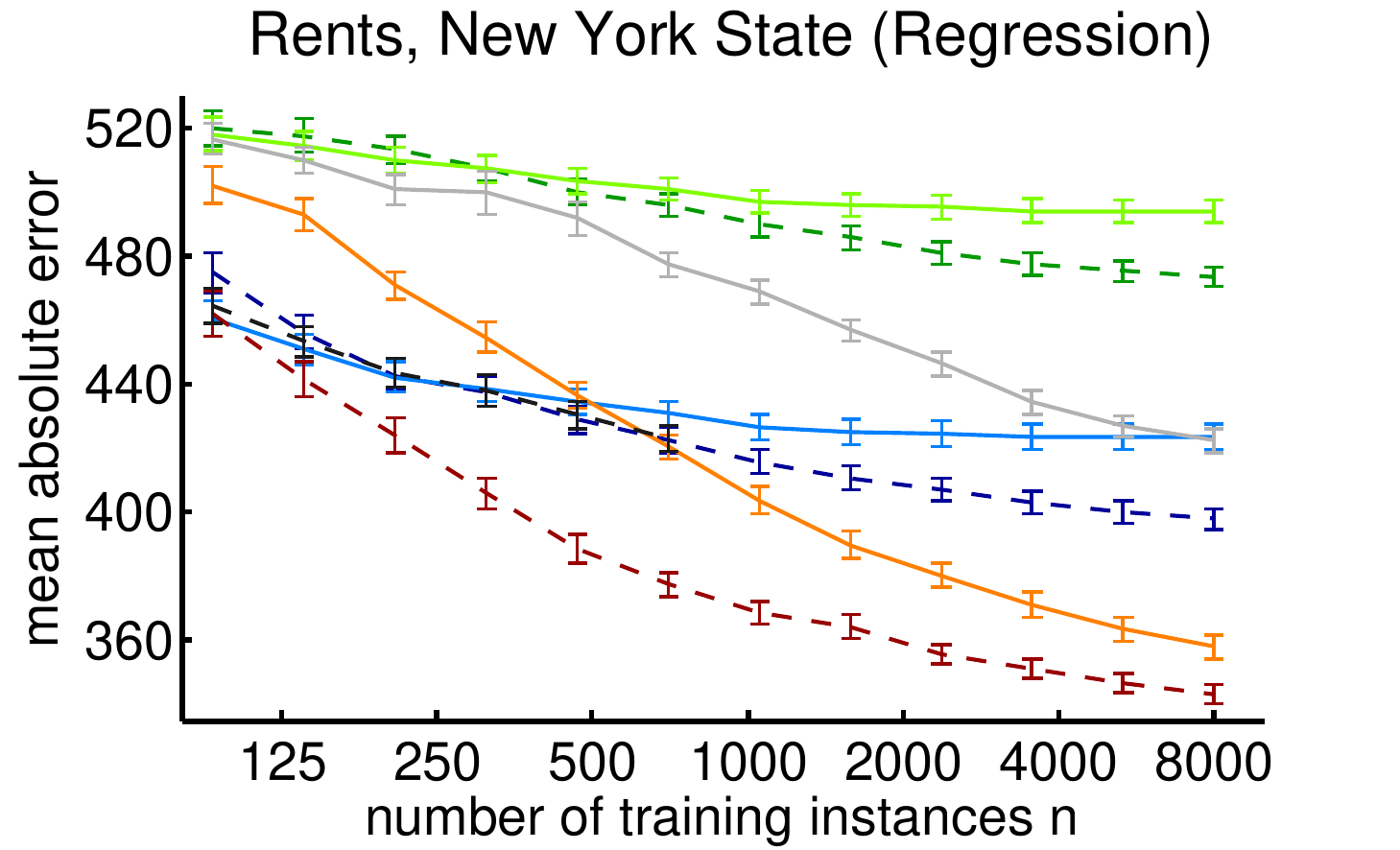}
		\hspace{1mm}
		\includegraphics[width=0.49\linewidth]{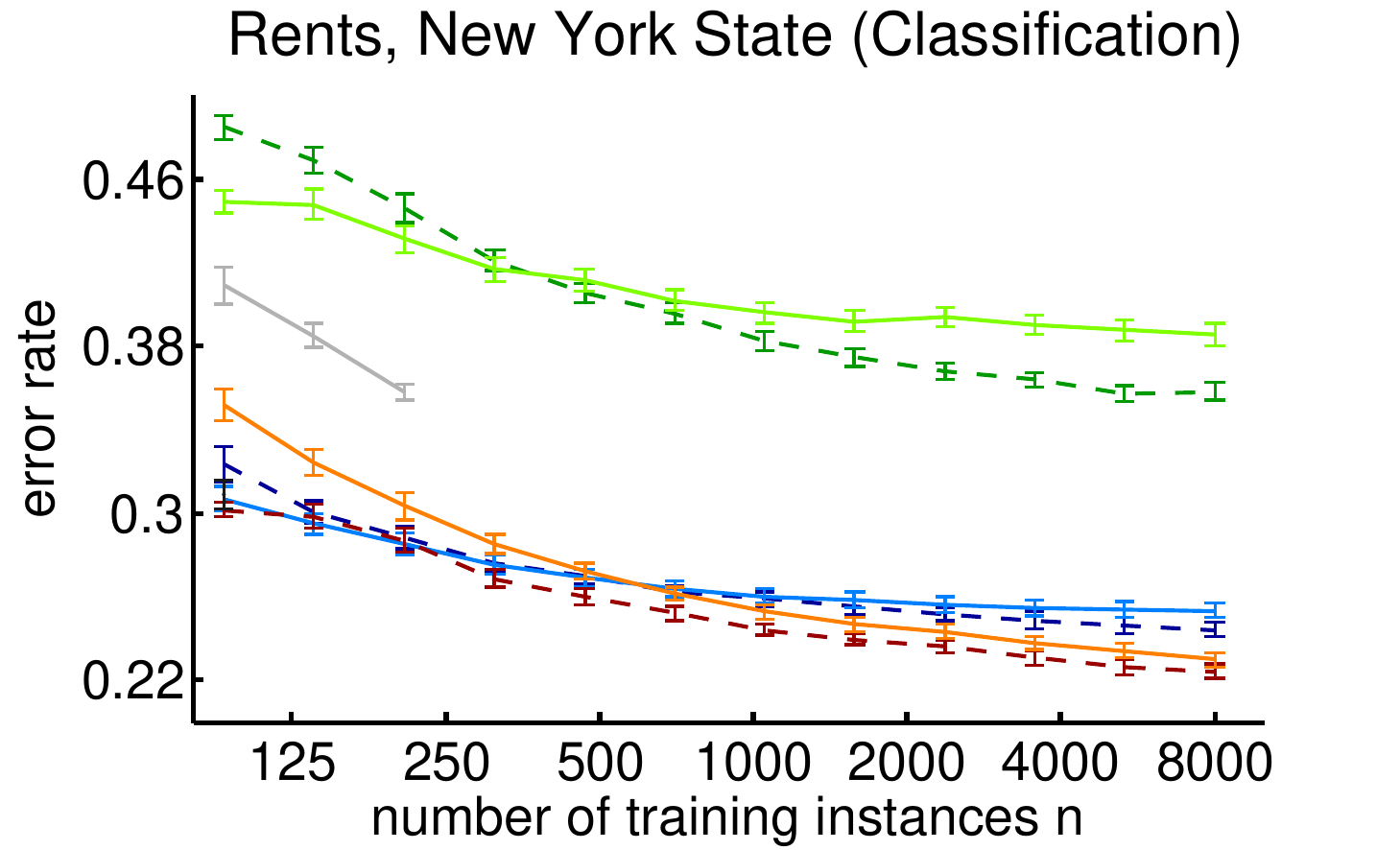}\\
		\vspace{5mm}
		\centering
		\fbox{\includegraphics[width=0.95\linewidth, viewport=0 52 560 72]{figures//LegendHorizontal}}
		\caption{
	Mean absolute error for predicting monthly housing rents (left) and mean zero-one loss for classifying rental contracts (right) in the states of California (upper row) and New York (lower row) over training set size $n$.		Error bars indicate the standard error.
			}
		\label{fig:main_results_rent}
	\end{center}
	\vskip -0.12in
\end{figure*}

Figure~\ref{fig:main_results} shows the mean absolute error for real estate price predictions (left) and the mean zero-one loss for classifying sales transactions (right) as a function of training set size $n$. 
For regression, \baselineFanLinear and \baselineFanRBF  partially completed the experiments; 
for classification, both methods did not complete the experiment for the smallest value of $n$. 
All other methods completed the experiments within the time limit.
For regression, we observe that \ourModelLinear is substantially more accurate than \baselineGPNaiveLinear, \baselineGPTimeGeoLinear, and \baselineFanLinear; \ourModelRBF is more accurate than \baselineGPNaiveRBF and \baselineGPTimeGeoRBF with $p<0.01$ for all training set sizes according to a paired $t$-test. Significance values of paired $t$-test comparing \ourModelRBF and \baselineFanRBF fluctuate between $p<0.01$ and $p < 0.2$ for different $n$, indicating that \ourModelRBF is likely more accurate than \baselineFanRBF.
For classification, 
\ourModelLinear substantially outperforms \baselineGPNaiveLinear and \baselineGPTimeGeoLinear; \ourModelRBF outperforms \baselineGPNaiveRBF and \baselineGPTimeGeoRBF ($p < 0.01$ for $n > 125$). 

Figure~\ref{fig:main_results_rent} shows the mean absolute error for predicting monthly housing rent (left) and the mean zero-one loss for classifying rental contracts (right) for rental contracts in the state of California (upper row) and the state of New York (lower row) as a function of training set size $n$. 
\baselineFanLinear completed the regression experiments within the time limit and partially completed
the classification experiment; \baselineFanRBF partially completed the regression experiment but did not complete the classification
experiment for the smallest value of $n$. 
We again observe that \ourModelRBF yields the most accurate predictions for both classification and regression problems; \ourModelLinear
always yields more accurate predictions than \baselineFanLinear and more accurate predictions than \baselineGPTimeGeoLinear for training set
sizes larger than $n = 1000$.

\section{Discussion and Related Work}
\label{sec:conc}

Varying-coefficient models reflect applications in which a conditional distribution of $y$ given $\bx$ is a function of task variables $\bt$. The task variables can, for instance, be continuous, discrete, or nodes in a tree---as in hierarchical Bayesian multitask learning. 
The functional dependency between the conditional distribution of the output given the input and the task variables can be modeled with a GP prior. 
Theorem~\ref{th:predDist} shows that, for isotropic GP priors, Bayesian inference in varying-coefficient models can be carried out efficiently by using a standard Gaussian process with a kernel that is defined as the product of a task kernel and an instance kernel. This result clarifies the exact modeling assumptions required to derive the multitask kernel of~\citet{bonilla2007kernel}. This result also highlights that Bayesian inference for hierarchical Bayesian learning can be carried out efficiently by using a standard Gaussian process with graph-Laplacian kernel \citep{evgeniou2005learning}. 

Product kernels play a role in other multitask learning models.
In the linear coregionalization model, several related functions are modeled as linear combinations
of Gaussian processes; the covariance function then resolves to a product of a kernel function 
on instances and a matrix of mixing coefficients~\citep{journel1978mining,alvarez2011kernels}.
A similar model is studied by~\citet{Wang2007multifactor} in the context of style-content separation
in human locomotion data; here mixing coefficients are given by latent variables
that represent an individual's movement style.
\citet{zhang2010convex} study a model for learning task relationships, and show that under a matrix-normal regularizer the 
solution of a multitask-regularized risk minimization problem can be expressed using a product kernel. 
Theorem~\ref{th:predDist} can be seen as a generalization of their result in which the regularizer is replaced by a prior over functions,
and the regularized risk minimization perspective by a fully Bayesian analysis.

Non-stationarity can also be modeled in Gaussian processes by assuming that either the residual variance\mbox{~\citep{wang2012gaussian}},
or the length scale of the covariance function~\citep{schmidt2003bayesian}, 
or the amplitude of the output~\citep{adams2008gaussian} are input-dependent. The varying-coefficient model differs from these models in that the source of non-stationarity is
observed in the task variable.

In the domain of real estate price prediction, the dependency between property attributes and the market price changes continuously with geographical coordinates and time. We observe that primal Bayesian inference in varying-coefficient models with nonisotropic GP priors is all but impractical in this domain, while for isotropic GP priors, inference based on Theorem~\ref{th:predDist} is more efficient by several orders of magnitude. 
Empirically, we observe that the linear and kernelized \ourModel models predict real estate prices and housing rents more accurately over time and space than kernel-local smoothing varying-coefficient models, and are also more accurate than linear and kernelized models that append the task variables to the attribute vector or ignore the task variables. 

\acks{
We would like to thank J\"orn Malich and Ahmed Abdelwahab for their help in preparing the data sets of monthly housing rents.
We gratefully acknowledge support from the German Research Foundation (DFG), grant LA 3270/1-1.
}

\appendix
\section*{Appendix}

\begin{proof}\textbf{of Proposition~\ref{prop:hb_drift}}.

\sloppy

The marginal $p(\by|\bX,\bT;\bkappa)$ is defined by the generative process of drawing
$\wfunc \sim \mathcal{GP}(\mathbf{0},\bkappa)$, evaluating $\wfunc$ for the $k$ different tasks to create parameter
vectors $\wfunc(1),\dots,\wfunc(k)$, and then drawing $y_i \sim p(y|\bx_i,\wfunc(\bt_i))$ for $i=1,\dots,n$.
The marginal $p(\by|\bX,\bT;\mathcal{G},\bsigma)$ is defined by the generative process of generating parameter vectors
$\wHB_1,\dots,\wHB_k$ according to 
Equations~\ref{eq:hb1} and~\ref{eq:hb2} 
in Definition~\ref{def:hb}, and then drawing 
$y_i \sim p(y|\bx_i,\wHB_{\bt_i})$ for $i=1,\dots,n$. Here, the observation models $p(y|\bx_i,\wHB_{\bt_i})$
and $p(y|\bx_i,\wfunc(\bt_i))$ are identical. It therefore suffices to show that
$p(\wfunc(1),\dots,\wfunc(k)|\bkappa) = p(\wHB_1,\dots,\wHB_k|\mathcal{G},\bsigma)$.

\fussy

The distribution $p(\wHB_1,\dots,\wHB_k|\mathcal{G},\bsigma)$ can be derived from standard results for Gaussian graphical 
models. 
Let $\bar{\bW} \in \mathbb{R}^{k \times m}$ denote the matrix with row vectors $\wHB_1 \trans,\dots,\wHB_k \trans$,
and let $\vect(\bar{\bW}\trans) \in \mathbb{R}^{km}$ denote the vector of random variables obtained by stacking the vectors $\wHB_1,\dots,\wHB_k$ on top of another. 
According to 
Equations~\ref{eq:hb1} and~\ref{eq:hb2}, 
the distribution
over the random variables within 
$\vect(\bar{\bW}\trans)$ 
is given by a Gaussian graphical model (\eg \cite{murphy2012machine}, Chapter 10.2.5) with weight matrix
$\bA\otimes \eyeMatrix{m} \in \mathbb{R}^{km \times km}$ and standard deviations $\bsigma \otimes \oneVector{m}$,
where $\oneVector{m}\in \mathbb{R}^{m}$ is the all-one vector. 
It follows that the distribution over 
$\vect(\bar{\bW}\trans) \in \mathbb{R}^{km}$ 
is given by
\begin{align*}
 p(\vect(\bar{\bW}\trans)|\mathcal{G},\bsigma) = 
\mathcal{N}(\vect(\bar{\bW}\trans)|\mathbf{0},\bar{\bSigma})
\end{align*}
with
\begin{align*}
\bar{\bSigma} = (\eyeMatrix{km}-\bA \otimes \eyeMatrix{m})^{-1}&\diag(\bsigma \otimes \oneVector{m})^2 \\
&(\eyeMatrix{km}-\bA\trans \otimes \eyeMatrix{m})^{-1}
\end{align*}
(see \cite{murphy2012machine}, Chapter 10.2.5),
where \mbox{$\diag(\bsigma \otimes \oneVector{m})\in \mathbb{R}^{km \times km}$} denotes the diagonal matrix with 
entries \mbox{$\bsigma \otimes \oneVector{m}$}.

The distribution $p(\wfunc(1),\dots,\wfunc(k)|\bkappa)$ is given directly by the Gaussian process defining the prior over vector-valued functions $\wfunc: \mathcal{T}\rightarrow \mathbb{R}^m$ (see Equation~\ref{eq:gp}). 
Let $\bOmega \in \mathbb{R}^{k \times m}$ denote the matrix with row vectors $\wfunc(1)\trans,\dots,\wfunc(k) \trans$,
then the Gaussian process prior implies
\begin{align*}
 p(\vect(\bOmega\trans)|\bkappa) = 
\mathcal{N}(\vect(\bOmega)\trans|\mathbf{0},\bG \otimes \eyeMatrix{m})
\end{align*}
(see, \eg \cite{alvarez2011kernels}, Section 3.3).
A straightforward calculation now shows $\bG \otimes \eyeMatrix{m} = \bar{\bSigma}$ and thereby proves the claim.
\end{proof}

\begin{proof}\textbf{of Proposition~\ref{prop:hb_laplace}}.
In the following we use the notation that is introduced in Proposition~\ref{prop:hb_drift} and Definition~\ref{def:graph_laplacian_kernel}. We first observe that by the definition of the graph Laplacian multitask kernel it is sufficient to show that $\bG = \bL^{\dagger}$. Since the matrix $\bG$ is invertible, this is equivalent to $\bG^{-1} = \bL$.

We prove the claim by induction over the number of nodes $|\mathcal{T}|$ in the tree $\mathcal{G}$. 
If $|\mathcal{T}| = 1$, then we have $\bA = 0$, $\bD = 0$, $\bR = \sigma_1^{-2}$ and $\bM = 0$. This leads to \[
\bG^{-1} = (\bI - \bA\trans)\sigma_1^{-2}(\bI - \bA)= \sigma_1^{-1} = \bD + \bR - \bM = \bL
\]
and proves the base case. Let us now assume that we have a tree $\mathcal{G}_k$ with $|\mathcal{T}| = k > 1$ nodes. Let $\bt$ be a leaf of this tree and $\bt'$ shall be its unique parent. Suppose we have $\bt' = i$ and w.l.o.g. we assume that $\bt = k$. Let furthermore $\mathcal{G}_{k-1}$ be the tree which we get by removing the node $k$ and its adjacent edge from the tree $\mathcal{G}_k$. 
Let $\bA_k$ and $\bA_{k-1}$ denote the adjacency matrices and $\bD_k$ and $\bD_{k-1}$ the degree matrices of $\mathcal{G}_k$ and $\mathcal{G}_{k-1}$.
Let $\bsigma_k \in \mathbb{R}^k$ be the vector with entries $\sigma_1,\dots,\sigma_k$, and $\bsigma_{k-1} \in \mathbb{R}^{k-1}$ be the vector with entries $\sigma_1,\dots,\sigma_{k-1}$. 
Let $\bR_k \in \mathbb{R}^{k \times k}$ denote the diagonal matrix with entries $\sigma_1^{-2},0,\dots,0$, and $\bR_{k-1} \in \mathbb{R}^{k-1 \times k-1}$
the diagonal matrix with entries $\sigma_1^{-2},0,\dots,0$. 
Let $\bB_k \in \mathbb{R}^{k \times k}$ denote the diagonal matrix with entries $0,\sigma_2^{-2},\dots,\sigma_k^{-2}$ and $\bB_{k-1} \in \mathbb{R}^{k-1 \times k-1}$ the diagonal matrix with entries $0,\sigma_2^{-2},\dots,\sigma_{k-1}^{-2}$.
Let $\bM_k = \bB_k \bA_k + (\bB_k \bA_k)\trans$ and $\bM_{k-1} = \bB_{k-1} \bA_{k-1} + (\bB_{k-1} \bA_{k-1})\trans$.
Let $\bL_k = \bD_k+\bR_k - \bM_k$ and $\bL_{k-1} = \bD_{k-1}+\bR_{k-1} - \bM_{k-1}$.

In the following, we write $\diag(\bv)$ to denote a diagonal matrix with entries $\bv$. 
We then have \[
\bA_k = \left( \begin{array}{c|c}
\bA_{k-1} & \be \\
\hline
\mathbf{0}& 0
\end{array} \right)\text{, where }\be = (\underbrace{0,\dots,0}_{i-1},1,0,\dots,0)\trans
\]
is the $i^{\text{th}}$ ($n-1$)-dimensional unit vector. Using this notation we can write
\begin{align*}
\bG_k^{-1} &= (\bI - \bA_k\trans)\diag(\bsigma_k)^{-2}(\bI - \bA_k)\\
&=\left(\begin{array}{c|c} \bI - \bA_{k-1}\trans & \mathbf{0} \\\hline-\be\trans& 1\end{array} \right)
\left(\begin{array}{c|c} \diag(\bsigma_{k-1})^{-2} & \mathbf{0} \\\hline\mathbf{0}& \sigma_{k}^{-2}\end{array} \right)
 \left(\begin{array}{c|c} \bI - \bA_{k-1} & -\be \\\hline\mathbf{0}& 1\end{array} \right)\\
&=\left(\begin{array}{c|c} \bL_{k-1} + \sigma_{k}^{-2}\be\be\trans & -\sigma_{k}^{-2}\be \\\hline-\sigma_{k}^{-2}\be\trans& \sigma_{k}^{-2}\end{array} \right).
\end{align*}
In the last line we applied the induction hypothesis to the tree $\mathcal{G}_{k-1}$. Using the definitions of $\bL$, $\bD$, 
$\bR$ and $\bM$, we can easily finish the proof:
\begin{align*}
\bG_{k}^{-1} &= \left(\begin{array}{c|c} \bD_{k-1} + \bR_{k-1} - \bM_{k-1}+ \sigma_{k}^{-2}\be\be\trans & -\sigma_{k}^{-2}\be \\\hline-\sigma_{k}^{-2}\be\trans& \sigma_{k}^{-2}\end{array} \right) \\
&=\left(\begin{array}{c|c} \bD_{k-1} + \sigma_{k}^{-2}\be\be\trans & \mathbf{0} \\\hline\mathbf{0}& \sigma_{k}^{-2}\end{array} \right)+
\left(\begin{array}{c|c} \bR_{k-1} & \mathbf{0} \\\hline\mathbf{0}& 0\end{array} \right)
-\left(\begin{array}{c|c} \bM_{k-1} & \sigma_{k}^{-2}\be \\\hline\sigma_{k}^{-2}\be\trans&0\end{array} \right)\\
 &= \bD_{k} + \bR_{k} - \bM_{k}\\
 &= \bL_{k}.
\end{align*}
This proves the claim.
\end{proof}

\bibliography{paper}

\end{document}